\documentclass[runningheads]{llncs}
\usepackage{graphicx}
\usepackage{cite}
\usepackage{algorithm,algorithmic}
\usepackage{booktabs}
\usepackage{multirow}
\usepackage{subcaption}
\usepackage{xcolor,colortbl}
\usepackage[hyphens]{url}
\usepackage{hyperref}
\usepackage{dsfont}
\usepackage{mathrsfs}

\usepackage[T1]{fontenc}
\usepackage{amsmath, amssymb}
\usepackage[ansinew]{inputenc}

\begin{document}
\title{FAIR-FATE: \\ Fair Federated Learning with Momentum}

\titlerunning{FAIR-FATE: \\ Fair Federated Learning with Momentum}

\author{Teresa Salazar\inst{1}\orcidID{0000-0003-2471-5783} \and
Miguel Fernandes\inst{1}\orcidID{0000-0002-7800-6854} \and
Helder Ara\'ujo\inst{2}\orcidID{0000-0002-9544-424X} \and
Pedro Henriques Abreu\inst{1}\orcidID{0000-0002-9278-8194}
}

\authorrunning{T. Salazar et al.}
\institute{Centre for Informatics and Systems, Department of Informatics Engineering of the University of Coimbra, University of Coimbra, Coimbra, Portugal \\
\email{\{tmsalazar,mfernandes,pha\}@dei.uc.pt}\\
\and
Institute of Systems and Robotics, Department of Electrical and Computer Engineering of the University of Coimbra, University of Coimbra, Coimbra, Portugal \\
\email{helder@isr.uc.pt}}

\maketitle              % typeset the header of the contribution

\begin{abstract}
While fairness-aware machine learning algorithms have been receiving increasing attention, the focus has been on centralized machine learning, leaving decentralized methods underexplored. Federated Learning is a decentralized form of machine learning where clients train local models with a server aggregating them to obtain a shared global model. Data heterogeneity amongst clients is a common characteristic of Federated Learning, which may induce or exacerbate discrimination of unprivileged groups defined by sensitive attributes such as race or gender. In this work we propose FAIR-FATE: a novel FAIR FederATEd Learning algorithm that aims to achieve group fairness while maintaining high utility via a fairness-aware aggregation method that computes the global model by taking into account the fairness of the clients. To achieve that, the global model update is computed by estimating a fair model update using a Momentum term that helps to overcome the oscillations of non-fair gradients. To the best of our knowledge, this is the first approach in machine learning that aims to achieve fairness using a fair Momentum estimate. Experimental results on real-world datasets demonstrate that FAIR-FATE outperforms state-of-the-art fair Federated Learning algorithms under different levels of data heterogeneity.

\footnote{This preprint has not undergone peer review or any post-submission improvements or corrections. The Version of Record of this contribution is published in ICCS 2023 - Lecture Notes in Computer Science, vol 14073, Springer, and is available online at \url{https://doi.org/10.1007/978-3-031-35995-8_37}}

\keywords{Fairness \and Federated Learning \and Machine Learning \and Momentum}
\end{abstract}

\section{Introduction}

With the widespread use of machine learning algorithms to make decisions which impact people's lives, the area of fairness-aware machine learning has been receiving increasing attention. Fairness-aware machine learning algorithms ensure that predictions do not prejudice unprivileged groups of the population with respect to sensitive attributes such as race or gender \cite{survey_fairness_ml}. However, the focus has been on centralized machine learning, with decentralized methods receiving little attention.

Federated Learning is an emerging decentralized technology that creates machine learning models using data distributed across multiple clients \cite{Federated_Learning_Avg}. In Federated Learning, the training process is divided into multiple clients which train individual local models on local datasets without the need to share private data. At each communication round, each client shares its local model updates with the server that uses them to create a shared global model.

While centralized machine learning batches can typically be assumed to be IID (independent and identically distributed), this assumption is unlikely to be true in Federated Learning settings \cite{distributions-nonidd}. Since in Federated Learning each client has its own private dataset, it may not be representative of all the demographic groups and thus can lead to discrimination of unprivileged groups defined by sensitive attributes. To solve the issue of heterogeneous client distributions, Federated Learning solutions based on Momentum gradient descent have been proposed \cite{best-fl, accelerating_fl_momentum}, but none of these has focused on fairness in machine learning. Moreover, typical centralized fair machine learning solutions require centralized access to the sensitive attributes information, and, consequently, cannot be directly applied to Federated Learning without violating privacy constraints. Because of these reasons, finding Federated Learning algorithms which facilitate collaboration amongst clients to build fair machine learning models while preserving their data privacy is a great challenge.

Motivated by the importance and challenges of developing fair Federated Learning algorithms, we propose FAIR-FATE: a novel fairness-aware Federated Learning algorithm that aims to achieve group fairness while maintaining classification performance. FAIR-FATE uses a new fair aggregation method that computes the global model by taking into account the fairness of the clients. In order to achieve this, the server has a validation set that is used at each communication round to measure the fairness of the clients and the current global model. Afterwards, the fair Momentum update is computed using a fraction of the previous fair update, as well as the average of the clients' updates that have higher fairness than the current global model, giving higher importance to fairer models. Since Momentum-based solutions have been used to help overcoming oscillations of noisy gradients, we hypothesize that the use of a fairness-aware Momentum term can overcome the oscillations of non-fair gradients. Therefore, this work aims to address the following research question:

\begin{quote}\emph{Can a fairness-aware Momentum term be used to overcome the oscillations of non-fair gradients in Federated Learning?}\end{quote}

Several experiments are conducted on real world datasets \cite{compas, adult, law-school, dutch-census} with a range of non-identical data distributions at each client, considering the different sensitive attributes and target classes in each dataset. Experimental results demonstrate that FAIR-FATE outperforms state-of-the-art fair Federated Learning algorithms on fairness without neglecting model performance under different data distributions.

\section{Related Work}

Fairness-aware machine learning has been gaining increasing attention. Fairness-aware algorithms have been categorized into three different groups, according to the stage in which they are performed: pre-processing, in-processing and post-processing \cite{survey_fairness_ml, best-fairness}. The algorithm proposed in this work falls into the in-processing category. Fairness metrics can be divided into two main groups: group fairness and individual fairness \cite{survey_fairness_ml}. Group fairness states that privileged and unprivileged groups should have equal probability of outcomes. On the other hand, individual fairness aims to give similar predictions to similar individuals. In this work, we focus on optimizing group fairness.

Few efforts have been made in the area of fairness-aware Federated Learning. Abay et al. \cite{mitigating-bias-fl} propose to apply three existing fair machine learning techniques to the Federated Learning setting. However, studies have shown that local debiasing strategies are ineffective in the presence of data heterogeneity \cite{parecido-fairl-fl-het-face}. Kanaparthy et al. \cite{parecido-fairl-fl-het-face} propose simple strategies for aggregating the local models based on different heuristics such as using the model which provides the least fairness loss for a given fairness notion or using the model with the highest ratio of accuracy with fairness loss. Ezzeldin et al. \cite{parecido-group-fairness-fl} propose a method to achieve global group fairness via adjusting the weights of different clients based on their local fairness metric. However, this work assumes that there is a local debiasing method at each client. Mehrabi et al. \cite{parecido-mehrabi} propose "FedVal" that is a simple aggregation strategy which performs a weighted averaged of the clients models based on scores according to fairness or performance measures on a validation set.

To the best of our knowledge, we have yet to see a fairness-aware machine learning approach that uses Momentum to estimate a fair update of the global model. Furthermore, FAIR-FATE is flexible since it does not rely on any fairness local strategy at each client, contrary to the works in \cite{parecido-group-fairness-fl, mitigating-bias-fl}. Finally, very few works consider multiple data heterogeneity scenarios (non-IID), which is a characteristic of Federated Learning settings.

\section{Problem Statement}

We first define the following notations used throughout the paper. We assume there exist $K$ clients in the Federated Learning setting, where each client has its private local dataset $D_k = \{\mathds{X}, Y \}, k \in K$. Each dataset $D_k$, contains $n_k$ instances where the total number of instances is defined as $\sum_{k=1}^{K} n_{k}$. We further assume a binary classification setting, where $\mathds{X} \in \mathcal{X}$ is the input space, $Y \in \mathcal{Y} = \{0, 1\}$ is the output space, $\hat{Y} \in \{0, 1\}$ represents the predicted class, and $S \in \{0, 1\}$ is the binary sensitive attribute of $\mathds{X}$.

\subsection{Federated Learning and Momentum}

The objective of Federated Learning is to train a global model, $\theta$, which resides on the server without access to the clients' private data. In Federated Learning, multiple clients collaborate to train a model that minimizes a weighted average of the loss across all clients. The objective of this Federated Learning framework can be written as \cite{Federated_Learning_Avg}:

\begin{equation}
    \min_{\theta} f(\theta) = \sum_{k \in K} \frac{n_{k}}{n} G_k(\theta), \qquad G_k(\theta) = \frac{1}{n_k} \sum_{i=1}^{n_k} f_{i}(\theta),
\label{eq:min}
\end{equation}
where $G_k(\theta)$ is the local objective for the client $k$ and $f_{i}(\theta)$ is the loss of the datapoint $i$ of client $k$ \cite{Federated_Learning_Avg}.

The first work on Federated Learning was proposed in \cite{Federated_Learning_Avg}, where the authors present an algorithm named Federated Averaging (FedAvg) that consists of a way of generating the global model by periodically averaging the clients' locally trained models. 

Since vanilla FedAvg can be slow to converge \cite{momentumgd}, Federated Learning solutions based on Momentum gradient descent have been proposed \cite{best-fl, accelerating_fl_momentum}. Momentum gradient descent is a technique where an exponentially weighted averaged of the gradients is used to help overcoming oscillations of noisy gradients. By providing a better estimate of the gradients, it is faster than stochastic gradient descent. Moreover, decaying Momentum rules have been previously applied to Federated Learning \cite{best-fl} with the objective of improving training and being less sensitive to parameter tuning by decaying the total contribution of a gradient to future updates.

\subsection{Fairness Metrics}

We consider the following group fairness metrics:

\begin{definition}[Statistical Parity (SP)] A predictor has Statistical Parity \cite{zafar-2017} if the proportion of individuals that were predicted to receive a positive output is equal for the two groups:

\begin{equation} \label{eq:sp}
    SP= \frac{P [\hat{Y} = 1 \mid S = 0]}{P [\hat{Y} = 1 \mid S = 1]}
\end{equation}

\end{definition}

\begin{definition}[Equality of Opportunity (EO)]
A classifier satisfies Equality of Opportunity (EO) if the True Positive Rate is the same across different groups \cite{equality-of-opportunity}:

\begin{equation}
    EO = \frac{P [\hat{Y} = 1 \mid S = 0, Y = 1]}{P [\hat{Y} = 1 \mid S = 1, Y = 1]}
\end{equation}

\end{definition}

\begin{definition}[Equalized Odds (EQO)]
A classifier satisfies Equalized Odds (EQO) if the True Positive Rate and False Positive Rate is the same across different groups \cite{equality-of-opportunity}:
\begin{equation}
    EQO= \frac{\frac{P [\hat{Y} = 1 \mid S = 0, Y = 0]}{P [\hat{Y} = 1 \mid S = 1, Y = 0]} + \frac{P [\hat{Y} = 1 \mid S = 0, Y = 1]}{P [\hat{Y} = 1 \mid S = 1, Y = 1]}}{2}
\end{equation}

\end{definition}

All of the metrics presented have an ideal value of 1. If a ratio is bigger than 1, $S=0$ becomes $S=1$ and $S=1$ becomes $S=0$.

\subsection{Objective} \label{objective}

The goal of our problem is to minimize the loss function $f(\theta)$ of the global model as well as to promote fairness without knowledge of the private data of the clients participating in the federation. We formulate a multi-objective optimization problem with two objectives: one for minimizing the model error, and one for promoting fairness:

\begin{align}
& \min_{\theta} \quad (1-\lambda_t) \underbrace{\sum_{k \in K_t} \frac{n_{k}}{n} G_k(\theta)}_\text{model error objective} + \lambda_t \underbrace{\sum_{k\in K_{tFair}} \frac{{F}_{\theta_{t+1}^k}}{{F}_{total}} G_k(\theta)}_\text{fairness promotion objective}
\end{align}

The model error objective is represented by the first term in the optimization problem, which involves minimizing the sum of the error functions over the clients participating in the federation round, $K_t$. This objective ensures that the resulting model performs well on the entire dataset. The fairness promotion objective is represented by the second term in the optimization problem, which involves minimizing the sum of the error functions over the clients of $K_t$ that have higher fairness values than the global model at the server, $K_{tFair}$, giving higher importance to clients which have higher values of fairness, $F \in \{SP, EO, EQO\}$. Hence, our objective does not only minimize the global loss to make the global model accurate, but also promotes fairness so that the final global model is both fair and accurate. To balance the trade-off between these two objectives we use $\lambda_t$ which is calculated at each federation round, $t$. The next section presents further details on how FAIR-FATE implements this objective.

\section{FAIR-FATE: Fair Federated Learning with Momentum} \label{section-fairfate}

FAIR-FATE is a novel fair Federated Learning algorithm developed in this work which consists of a new approach of aggregating the clients' local models in a fair manner, and, simultaneously, using a fairness-aware Momentum term that has the objective of overcoming the oscillations of non-fair gradients. 

In this approach, the server has access to a small validation set, similar to the works in \cite{parecido-mehrabi, validation-set}, which discuss different ways of obtaining this set. \footnote{For the sake of simplicity, we chose to randomly select a portion of the dataset as the validation set. To ensure the results are based on a representative sample, we conducted the experiments 10 times with a different validation set randomly selected each time.} Given an update from a client, the server can use the validation set to verify whether the client's update would improve the global model's fairness. 

The algorithm starts by initializing a global model $\theta_t$. Then, at each round, the server sends its current model, $\theta_{t}$, to a random subset of clients, $K_t$ of size ${\mid}K_t{\mid}$, ${\mid}K_t{\mid} < {\mid}K{\mid}$. The chosen clients then train their local models using their private data, producing an updated model $\theta_{t+1}^k$.

\subsubsection{Two model updates: $\alpha_N$ and $\alpha_F$}

To simultaneously achieve the two objectives described in Section \ref{objective}, the server calculates two model updates: $\alpha_N$, corresponding to the model error objective, and $\alpha_F$, corresponding to the fairness promotion objective. $\alpha_N$ is calculated by averaging the local updates of each client, given by:
\begin{equation}
    \alpha_N = \sum_{k \in K_t} \frac{n_k}{n} (\theta_{t+1}^k - \theta_{t}),
\end{equation}
where $\theta_{t}$ is the global model at communication round $t$, and $\theta_{t+1}^k$ is the local model of client $k$ at communication round $t + 1$. On the other hand, $\alpha_F$ is a weighted average of the clients' updates which have higher fairness than the current global model at the server on the validation set, given by:
\begin{equation}
    \alpha_F = \sum_{k \in K_{tFair}} \frac{{F}_{\theta_{t+1}^k}}{{F}_{total}} (\theta_{t+1}^k - \theta_{t}), \ \textnormal{where} \quad {F}_{total} = \sum_{k \in K_{tFair}} {F}_{\theta_{t+1}^k},
\end{equation}
As such, the weighted average gives higher importance to models with higher fairness values on the validation set.

\subsubsection{Decaying Momentum}

Afterwards we use Momentum, and, more specifically, decaying Momentum \cite{best-fl, chen2022demon}, calculating $\beta_t$ as follows:
\begin{equation}
    \beta_{t} = \beta_0 \times \frac{(1 - \frac{t}{T})}{(1 - \beta_0) + \beta_0(1 - \frac{t}{T})},
\end{equation}
where $\beta_0$ is the initial Momentum parameter. The fraction $(1 - \frac{t}{T})$ refers to the proportion of iterations remaining. The Momentum update, $v_{t+1}$, is calculated by summing a fraction of the previous update, $v_{t}$, and the global model fair update, $\alpha_F$, as follows:

\begin{equation}
    v_{t+1} = \beta_t v_{t} + (1 - \beta_t) \alpha_F,
\end{equation}
where $\beta_t$ controls the amount of the global fair update against the previous update.

The use of decaying Momentum is motivated by two main reasons. Firstly, when $\alpha_F$ is calculated using the local updates of the clients which have higher fairness than the server, it can be affected by the variability and noise of the data of each individual client. This can result in oscillations and instability in the overall training process. By introducing Momentum, we can smooth out these oscillations and stabilize the overall training process, leading to better convergence of the overall optimization problem. Secondly, using decaying Momentum can take advantage of momentum for speed-up in the early phases of training, but decay Momentum throughout training \cite{best-fl, chen2022demon}. This can prevent neural network weights from growing excessively fast, which has been found to be crucial for achieving optimal performance \cite{heo2020adamp}.

\subsubsection{Exponential Growth}

Finally, the global model, $\theta_{t+1}$, is updated by summing the previous model, $\theta_{t}$, with the fair Momentum update, $v_{t+1}$, and the normal update, $\alpha_N$, as follows:

\begin{equation}
    \theta_{t+1} = \theta_{t} + \lambda_t v_{t+1} + (1 - \lambda_t) \alpha_N,
\end{equation}
where $\lambda_t$ has the objective of controlling the amount of estimated fairness against the actual model update. It follows an exponential growth calculated as follows:

\begin{equation}
     \lambda_t = \lambda_0 (1 + \rho)^t , \qquad \lambda_t \leq MAX,
\end{equation}
where $\rho$, $\lambda_0$ and $MAX$ are hyperparameters that can be tuned which represent the growth rate, the initial amount, and the upper bound of $\lambda_t$, respectively. 

By exponentially increasing $\lambda$, FAIR-FATE allows that in the early stages of training, the model will prioritize the optimization of the main objective function using $\alpha_N$. In the latter stages, when it is expected that the model has a good performance and there is a better Momentum estimate, FAIR-FATE will have a higher value of $\lambda$ and, as a consequence, it will prioritize the global model's fair update, $\alpha_F$. Regarding the $MAX$ hyperparameter, a higher value will prioritize fair model updates, while a lower value will take more the model performance into consideration. Hence, as it will be observed in the experimental section, the $MAX$ hyperparameter can be used to control the fairness and model performance trade-off. Algorithm \ref{algo:fair-fate} presents the FAIR-FATE algorithm.

\begin{algorithm}[h!]
\caption{FAIR-FATE}
\label{algo:fair-fate}
 \begin{algorithmic}[1]
     \STATE Initialize global model: $\theta_t$, number of rounds: $T$, growth rate: $\rho$, upper bound of $\lambda_t$: $MAX$, Momentum parameter: $\beta_t$, initial Momentum parameter: $\beta_0$, initial amount: $\lambda_0$, initial Momentum update: $v_{0} = 0$.
     \FOR{each round $t$ = 1,2,..., $T$}
        \STATE Select $K_t$ = random subset of clients $K$
        \FOR{each client $k \in K_{t}$}
            \STATE Send $\theta_{t}$ and receive the locally trained model, $\theta_{t+1}^k$
        \ENDFOR
        \STATE $K_{tFair} \in K_t,$  \quad where ${F}_{\theta_{t+1}^{k}} >= {F}_{\theta_{t}}$
        \STATE ${F}_{total} = \sum_{k \in K_{tFair}} {F}_{\theta_{t+1}^k}$
        \STATE $\alpha_F = \sum_{k \in K_{tFair}} \frac{{F}_{\theta_{t+1}^k}}{{F}_{total}} (\theta_{t+1}^k - \theta_{t})$
        \STATE $\alpha_N = \sum_{k \in K_t} \frac{n_k}{n} (\theta_{t+1}^k - \theta_{t})$
        \STATE $\beta_{t} = \beta_0 \times \frac{(1 - \frac{t}{T})}{(1 - \beta_0) + \beta_0(1 - \frac{t}{T})}$
        \STATE $v_{t+1} = \beta_t v_{t} + (1 - \beta_t) \alpha_F$
        \STATE $\lambda_t = \lambda_0 (1 + \rho)^t , \qquad \lambda_t \leq MAX$
        \STATE $\theta_{t+1} = \theta_{t} + \lambda_t v_{t+1} + (1 - \lambda_t) \alpha_N$
     \ENDFOR
 \end{algorithmic}
\end{algorithm}

\section{Experiments}

In this section, a comprehensive set of experiments to evaluate FAIR-FATE is conducted in a Federated Learning scenario.

\subsection{Datasets}
FAIR-FATE is evaluated on four real-world datasets which have been widely used in the literature \cite{survey_fairness_ml} to access the performance of fairness-aware algorithms: the \texttt{COMPAS} \cite{compas}, the \texttt{Adult} \cite{adult}, the \texttt{Law School} \cite{law-school}, and the \texttt{Dutch} \cite{dutch-census} datasets. The \texttt{COMPAS} dataset's prediction task is to calculate the recidivism outcome, indicating whether individuals will be rearrested within two years after the first arrest, with race as the sensitive attribute. The \texttt{Adult} dataset's prediction task is to determine whether a person's income exceeds \$50K/yr or not, with gender as the sensitive attribute. The \texttt{Law School} dataset's prediction task is to predict whether a candidate would pass the bar exam, with race as the sensitive attribute. Finally, the \texttt{Dutch} Census dataset's prediction task is to determine a person's occupation which can be categorized as high level (prestigious) or low level profession, with gender as the sensitive attribute.

\subsection{Non-identical Client Data}

Because of its decentralized nature, Federated Learning exacerbates the problem of bias since clients' data distributions can be very heterogeneous. This means that there may exist a client that contains a very high representation of datapoints belonging to specific sensitive group with a certain outcome (e.g. unprivileged group with positive outcome - $S=0, Y=1$) and another client that has a very low representation of the same group.

Following the works in \cite{distributions-nonidd, parecido-group-fairness-fl}, we study the algorithms under different sensitive attribute distributions across clients. We use a non-IID synthesis method based on the Dirichlet distribution proposed in \cite{distributions-nonidd} and apply it in configurable sensitive attribute distribution such as the work in \cite{parecido-group-fairness-fl} but including the target distributions as well. More specifically, for each sensitive attribute value $s$ and target value $y$, we sample $p_{s,y} \sim Dir(\sigma)$ and allocate a portion $p_{s,y,k}$ of the datapoints with $S = s$ and $Y = y$ to client $k$. The parameter $\sigma$ controls the heterogeneity of the distributions in each client, where $\sigma \longrightarrow \infty$ results in IID distributions \cite{parecido-group-fairness-fl}.

\begin{figure}
    \centering
    \subfloat[\centering $\sigma=0.5$]{{\includegraphics[width=5.75cm]{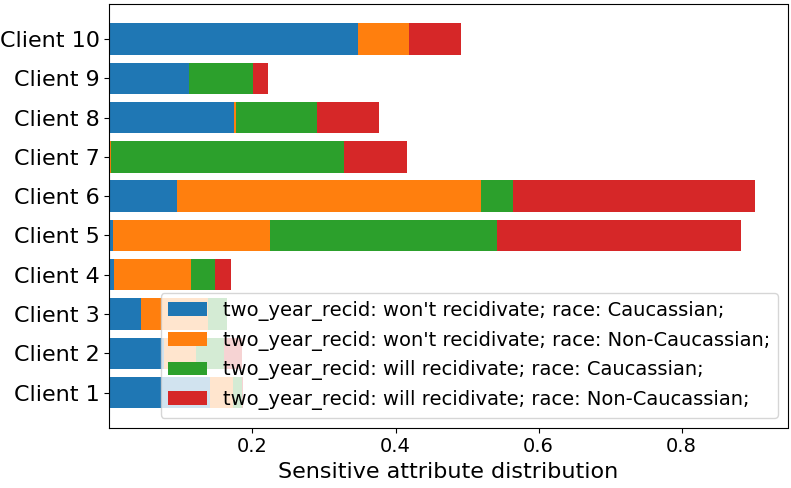}}}%
    \qquad
    \subfloat[\centering $\sigma \rightarrow{} \infty$]{{\includegraphics[width=5.75cm]{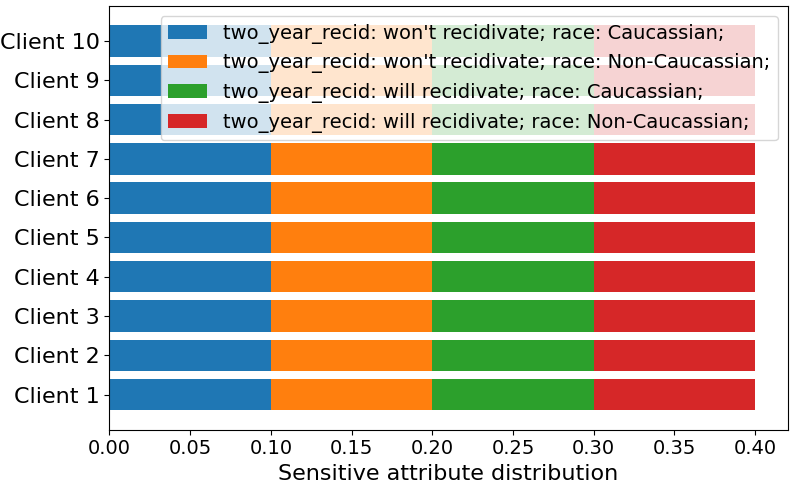}}}%
    \caption{Examples of heterogeneous data distributions on the \texttt{COMPAS} dataset using race as the sensitive feature for 10 clients with $\sigma=0.5$ and $\sigma \rightarrow{} \infty$}%
    \label{fig:distributions}%
\end{figure}

Figure \ref{fig:distributions} presents different examples of heterogeneous data distributions on the \texttt{COMPAS} dataset using race as the sensitive feature for 10 clients with $\sigma=0.5$ and $\sigma \rightarrow{} \infty$. On one hand it can be observed that for $\sigma=0.5$, different clients have very different representations of privileged and unprivileged groups and positive and negative outcomes. For example, while client 10 has about 40\% of the Caucasians that won't recidivate, client 9 has only about 10\%. Moreover, it can be observed that different clients have different number of datapoints. On the other hand, for $\sigma \rightarrow{} \infty$, the representations of the different groups for different clients are the same.

\subsection{Implementation and Setup Details}

We train a fully connected feed-forward neural network for binary classification on the described datasets with one hidden layer with 10 $tanh$ units. On the local clients models, mini-batch stochastic gradient descent and the standard binary cross-entropy loss are used.

The data instances are split into 60\% for the training set, 20\% for the server's validation set and 20\% for the testing set. The training data is divided into private training datasets, one for each client. The private training datasets owned by the clients are used to train the local models. The validation set is used to calculate the fairness of the models for the fair Momentum update as explained in the previous sections. The testing dataset is used to evaluate the model performance and fairness of the proposed approach and the baselines. The presented results are based on the average values over 10 runs.

We set $\lambda_0 = \{0.1, 0.5\}$, $\rho =\{0.04, 0.05\}$, $MAX = \{0.8, 0.9, 1.0\}$, $\beta_0 = \{0.8, 0.9, 0.99\}$, $F = \{SP, EO, EQO\}$. Additionally, we set $\sigma = \{0.5, 1.0\}$, $T = 100$, $E = 10$ (local epochs), $B = 10$ (local batches), $\eta_l=0.01$ (local learning rate). Based on the number of instances of the \texttt{COMPAS}, the \texttt{Adult}, the \texttt{Law School} and the \texttt{Dutch} datasets, we set ${\mid}K{\mid}$ to 10, 15, 12 and 20, respectively, and ${\mid}K_t{\mid}$ to about a third of ${\mid}K{\mid}$: 3, 5, 4, 6, respectively. \footnote{Source code can be found at: \url{https://github.com/teresalazar13/FAIR-FATE}.}

The following algorithms are used as the baselines:

\begin{itemize}
    \item FedAvg - the classical Federated Averaging algorithm presented in \cite{Federated_Learning_Avg};
    \item FedMom - the Federated Averaging with Momentum presented in \cite{accelerating_fl_momentum};
    \item FedDemon - the Federated Averaging with Momentum Decay presented \cite{best-fl};
    \item FedAvg+LR - the fair Federating Learning approach proposed in \cite{mitigating-bias-fl} where each client uses a local reweighing technique;
    \item FedAvg+GR - the fair Federating Learning approach proposed in \cite{mitigating-bias-fl} where each client uses a global reweighing technique;
    \item FedVal - the fair Federating Learning approach proposed in \cite{parecido-mehrabi} where the server performs a weighted average based on the validation scores. 
\end{itemize}

\subsection{Experimental Results}

\begin{table}[h!]
\caption{Results for the \texttt{COMPAS} and \texttt{Adult} datasets under different $\sigma$ heterogeneity levels and random (RND) splits. \newline FedAvg (1), FedMom (2), FedDemon (3), FedAvg+LR (4), FedAvg+GR (5), FedVal ($F=SP$), (6) FedVal, ($F=EO$) (7), FedVal ($F=EQO$) (8), FAIR-FATE ($F=SP$) \textbf{(9)}, FAIR-FATE ($F=EO$) \textbf{(10)}, FAIR-FATE ($F=EQO$) \textbf{(11)}} \label{table:res}
\centering
\begin{tabular}{c|c|c|c|c|c|c|c} \toprule
     \multirow{3}{*}{} & \multirow{3}{*}{Alg.} & \multicolumn{3}{c|}{\texttt{COMPAS} (race)} & \multicolumn{3}{c}{\texttt{Adult} (gender)} \\
     \cline{3-8}
     & & \multicolumn{2}{c|}{$\sigma$} & \multirow{2}{*}{RND} & \multicolumn{2}{c|}{$\sigma$} & \multirow{2}{*}{RND} \\
     \cline{3-4} \cline{6-7}
     & & 0.5 & 1.0 & & 0.5 & 1.0 & \\ \toprule
     \multirow{11}{*}{\rotatebox[origin=c]{90}{\textit{ACC}}} 
     & (1)              & 0.60+-0.04 & 0.64+-0.03 & 0.68+-0.01 & 0.79+-0.05 & 0.83+-0.01 & 0.84+-0.01 \\ 
     & (2)              & 0.62+-0.05 & 0.63+-0.04 & 0.69+-0.01 & 0.83+-0.02 & 0.82+-0.01 & 0.82+-0.01 \\ 
     & (3)              & 0.68+-0.01 & 0.68+-0.01 & 0.68+-0.01 & 0.84+-0.01 & 0.84+-0.01 & 0.84+-0.01 \\
     & (4)              & 0.62+-0.04 & 0.65+-0.02 & 0.68+-0.01 & 0.81+-0.03 & 0.82+-0.01 & 0.84+-0.01 \\  
     & (5)              & 0.62+-0.04 & 0.65+-0.03 & 0.68+-0.01 & 0.80+-0.03 & 0.82+-0.01 & 0.84+-0.01 \\
     & (6)              & 0.61+-0.04 & 0.64+-0.03 & 0.68+-0.01 & 0.75+-0.10 & 0.83+-0.01 & 0.84+-0.01 \\
     & (7)              & 0.62+-0.05 & 0.66+-0.03 & 0.68+-0.01 & 0.78+-0.06 & 0.82+-0.03 & 0.84+-0.01 \\
     & (8)              & 0.62+-0.04 & 0.66+-0.03 & 0.68+-0.01 & 0.76+-0.07 & 0.82+-0.03 & 0.84+-0.01 \\
     & \textbf{(9)}     & 0.57+-0.02 & 0.61+-0.04 & 0.66+-0.01 & 0.74+-0.14 & 0.80+-0.06 & 0.80+-0.02 \\  
     & \textbf{(10)}    & 0.63+-0.04 & 0.65+-0.02 & 0.67+-0.01 & 0.74+-0.09 & 0.78+-0.06 & 0.83+-0.02 \\  
     & \textbf{(11)}    & 0.63+-0.04 & 0.65+-0.03 & 0.67+-0.01 & 0.75+-0.09 & 0.79+-0.05 & 0.77+-0.01 \\ \midrule
     
     \multirow{7}{*}{\rotatebox[origin=c]{90}{\textbf{\textit{SP}}}} 
     & (1)              & 0.59+-0.34 & 0.86+-0.17 & 0.83+-0.04 & 0.41+-0.30 & 0.27+-0.10 & 0.31+-0.02 \\ 
     & (2)              & 0.75+-0.23 & 0.83+-0.20 & 0.82+-0.03 & 0.39+-0.16 & 0.63+-0.24 & 0.46+-0.20 \\
     & (3)              & 0.83+-0.04 & 0.83+-0.04 & 0.83+-0.04 & 0.31+-0.03 & 0.31+-0.03 & 0.31+-0.03 \\
     & (4)              & 0.93+-0.10 & 0.91+-0.08 & 0.94+-0.02 & 0.69+-0.19 & 0.70+-0.10 & 0.67+-0.05 \\  
     & (5)              & 0.76+-0.28 & 0.87+-0.16 & 0.94+-0.02 & 0.66+-0.27 & 0.64+-0.20 & 0.67+-0.05 \\
     & (6)              & 0.87+-0.17 & 0.88+-0.13 & 0.83+-0.04 & 0.71+-0.20 & 0.52+-0.18 & 0.32+-0.02 \\
     & \textbf{(9)}     & \cellcolor{blue!14}{\textbf{1.00+-0.01}} & \cellcolor{blue!14}{\textbf{0.98+-0.01}} & \cellcolor{blue!14}{\textbf{0.95+-0.03}} & \cellcolor{blue!14}{\textbf{0.79+-0.16}} & \cellcolor{blue!14}{\textbf{0.81+-0.11}} & \cellcolor{blue!14}{\textbf{0.79+-0.13}} \\ 
     & & \multicolumn{1}{r|}{\cellcolor{blue!14}{\textbf{+7\%}}} & \multicolumn{1}{r|}{\cellcolor{blue!14}{\textbf{+7\%}}} & \multicolumn{1}{r|}{\cellcolor{blue!14}{\textbf{+1\%}}} & \multicolumn{1}{r|}{\cellcolor{blue!14}{\textbf{+8\%}}} & \multicolumn{1}{r|}{\cellcolor{blue!14}{\textbf{+11\%}}} & \multicolumn{1}{r}{\cellcolor{blue!14}{\textbf{+8\%}}} \\ \midrule
     
     \multirow{7}{*}{\rotatebox[origin=c]{90}{\textbf{\textit{EO}}}} 
     & (1)              & 0.46+-0.34 & 0.81+-0.19 & 0.85+-0.05 & 0.51+-0.27 & 0.68+-0.21 & 0.78+-0.04 \\ 
     & (2)              & 0.74+-0.24 & 0.73+-0.34 & 0.83+-0.05 & 0.66+-0.26 & 0.65+-0.17 & 0.77+-0.17 \\
     & (3)              & 0.84+-0.05 & 0.84+-0.05 & 0.84+-0.05 & 0.78+-0.04 & 0.78+-0.04 & 0.78+-0.04 \\
     & (4)              & 0.77+-0.31 & 0.91+-0.08 & \cellcolor{red!14}{\textbf{0.98+-0.03}} & 0.66+-0.21 & 0.69+-0.15 & 0.76+-0.03 \\  
     & (5)              & 0.54+-0.38 & 0.84+-0.21 & \cellcolor{red!14}{\textbf{0.97+-0.03}} & 0.55+-0.28 & 0.71+-0.14 & 0.75+-0.04 \\
     & (7)              & 0.58+-0.34 & 0.83+-0.22 & 0.85+-0.05 & 0.72+-0.15 & 0.80+-0.12 & 0.79+-0.04 \\
     & \textbf{(10)}    & \cellcolor{blue!14}{\textbf{0.95+-0.06}} & \cellcolor{blue!14}{\textbf{0.96+-0.02}} & 0.95+-0.03 & \cellcolor{blue!14}{\textbf{0.85+-0.15}} & \cellcolor{blue!14}{\textbf{0.95+-0.02}} & \cellcolor{blue!14}{\textbf{0.90+-0.07}} \\ 
     & & \multicolumn{1}{r|}{\cellcolor{blue!14}{\textbf{+11\%}}} & \multicolumn{1}{r|}{\cellcolor{blue!14}{\textbf{+5\%}}} & \multicolumn{1}{r|}{\cellcolor{red!14}{\textbf{-3\%}}} & \multicolumn{1}{r|}{\cellcolor{blue!14}{\textbf{+7\%}}} & \multicolumn{1}{r|}{\cellcolor{blue!14}{\textbf{+15\%}}} & \multicolumn{1}{r}{\cellcolor{blue!14}{\textbf{+11\%}}} \\ \midrule
     
     \multirow{7}{*}{\rotatebox[origin=c]{90}{\textbf{\textit{EQO}}}} 
     & (1)                           & 0.39+-0.32 & 0.77+-0.19 & 0.76+-0.06 & 0.36+-0.21 & 0.42+-0.15 & 0.48+-0.03 \\ 
     & (2)                           & 0.65+-0.26 & 0.71+-0.26 & 0.73+-0.05 & 0.47+-0.24 & 0.53+-0.21 & 0.60+-0.19 \\
     & (3)                         & 0.75+-0.06 & 0.75+-0.06 & 0.75+-0.06 & 0.48+-0.04 & 0.48+-0.04 & 0.48+-0.04 \\
     & (4)                        & 0.78+-0.29 & 0.87+-0.11 & \cellcolor{red!14}{\textbf{0.93+-0.03}} & 0.58+-0.23 & 0.75+-0.09 & 0.80+-0.05 \\  
     & (5)                        & 0.59+-0.26 & 0.82+-0.21 & \cellcolor{red!14}{\textbf{0.93+-0.04}} & 0.46+-0.30 & 0.69+-0.19 & 0.80+-0.05 \\
     & (8)                 & 0.62+-0.29 & 0.83+-0.17 & 0.76+-0.06 & 0.59+-0.21 & 0.56+-0.12 & 0.49+-0.03 \\
     & \textbf{(11)}     & \cellcolor{blue!14}{\textbf{0.94+-0.05}} & \cellcolor{blue!14}{\textbf{0.93+-0.03}} & 0.92+-0.04 & \cellcolor{blue!14}{\textbf{0.78+-0.10}} & \cellcolor{blue!14}{\textbf{0.79+-0.11}} & \cellcolor{blue!14}{\textbf{0.82+-0.10}} \\ 
     & & \multicolumn{1}{r|}{\cellcolor{blue!14}{\textbf{+16\%}}} & \multicolumn{1}{r|}{\cellcolor{blue!14}{\textbf{+6\%}}} & \multicolumn{1}{r|}{\cellcolor{red!14}{\textbf{-1\%}}} &  \multicolumn{1}{r|}{\cellcolor{blue!14}{\textbf{+19\%}}} & \multicolumn{1}{r|}{\cellcolor{blue!14}{\textbf{+4\%}}} & \multicolumn{1}{r}{\cellcolor{blue!14}{\textbf{+2\%}}} \\ \bottomrule
\end{tabular}
\end{table}

\begin{table}[h!]
\caption{Results for the \texttt{Law School} and \texttt{Dutch} Census datasets under different $\sigma$ heterogeneity levels and random (RND) splits. \newline FedAvg (1), FedMom (2), FedDemon (3), FedAvg+LR (4), FedAvg+GR (5), FedVal ($F=SP$), (6) FedVal, ($F=EO$) (7), FedVal ($F=EQO$) (8), FAIR-FATE ($F=SP$) \textbf{(9)}, FAIR-FATE ($F=EO$) \textbf{(10)}, FAIR-FATE ($F=EQO$) \textbf{(11)}} \label{table:res-2}
\centering
\begin{tabular}{c|c|c|c|c|c|c|c} \toprule
     \multirow{3}{*}{} & \multirow{3}{*}{Alg.} & \multicolumn{3}{c|}{\texttt{Law School} (race)} & \multicolumn{3}{c}{\texttt{Dutch} Census (gender)} \\
     \cline{3-8}
     & & \multicolumn{2}{c|}{$\sigma$} & \multirow{2}{*}{RND} & \multicolumn{2}{c|}{$\sigma$} & \multirow{2}{*}{RND} \\
     \cline{3-4} \cline{6-7}
     & & 0.5 & 1.0 & & 0.5 & 1.0 & \\ \toprule
     \multirow{11}{*}{\rotatebox[origin=c]{90}{\textit{ACC}}} 
     & (1)               & 0.89+-0.01 & 0.89+-0.02 & 0.90+-0.01 & 0.81+-0.02 & 0.82+-0.01 & 0.83+-0.00 \\ 
     & (2)               & 0.89+-0.01 & 0.90+-0.01 & 0.90+-0.01 & 0.81+-0.01 & 0.82+-0.01 & 0.83+-0.00 \\ 
     & (3)               & 0.90+-0.01 & 0.90+-0.01 & 0.90+-0.01 & 0.82+-0.01 & 0.82+-0.01 & 0.82+-0.01 \\
     & (4)               & 0.89+-0.00 & 0.88+-0.03 & 0.89+-0.01 & 0.81+-0.03 & 0.82+-0.01 & 0.82+-0.03 \\  
     & (5)               & 0.90+-0.01 & 0.88+-0.01 & 0.89+-0.01 & 0.81+-0.02 & 0.82+-0.01 & 0.82+-0.00 \\
     & (6)               & 0.88+-0.03 & 0.89+-0.01 & 0.90+-0.01 & 0.78+-0.04 & 0.82+-0.01 & 0.83+-0.00 \\
     & (7)               & 0.80+-0.12 & 0.88+-0.02 & 0.90+-0.00 & 0.81+-0.02 & 0.82+-0.01 & 0.83+-0.00 \\
     & (8)               & 0.85+-0.07 & 0.89+-0.02 & 0.99+-0.00 & 0.80+-0.03 & 0.82+-0.01 & 0.83+-0.00 \\
     & \textbf{(9)}      & 0.89+-0.00 & 0.89+-0.01 & 0.89+-0.01 & 0.75+-0.07 & 0.78+-0.04 & 0.81+-0.01 \\  
     & \textbf{(10)}     & 0.90+-0.01 & 0.89+-0.03 & 0.90+-0.00 & 0.77+-0.07 & 0.77+-0.08 & 0.82+-0.00 \\  
     & \textbf{(11)}     & 0.90+-0.01 & 0.90+-0.01 & 0.90+-0.01 & 0.75+-0.08 & 0.76+-0.06 & 0.81+-0.01 \\ \midrule
     
     \multirow{7}{*}{\rotatebox[origin=c]{90}{\textbf{\textit{SP}}}} 
     & (1)               & 0.87+-0.13 & 0.88+-0.09 & 0.89+-0.02 & 0.60+-0.12 & 0.61+-0.09 & 0.60+-0.02 \\ 
     & (2)               & 0.91+-0.12 & 0.91+-0.07 & 0.91+-0.04 & 0.62+-0.09 & 0.59+-0.10 & 0.60+-0.01 \\
     & (3)               & 0.91+-0.03 & 0.91+-0.03 & 0.91+-0.03 & 0.60+-0.01 & 0.60+-0.01 & 0.60+-0.01 \\
     & (4)               & 0.96+-0.04 & 0.95+-0.06 & 0.98+-0.01 & 0.72+-0.04 & 0.72+-0.04 & 0.74+-0.02 \\  
     & (5)               & 0.95+-0.05 & 0.97+-0.03 & 0.98+-0.01 & 0.72+-0.10 & 0.73+-0.08 & 0.74+-0.02 \\
     & (6)               & 0.89+-0.11 & 0.92+-0.05 & 0.89+-0.02 & 0.67+-0.13 & 0.70+-0.09 & 0.60+-0.01 \\
     & \textbf{(9)}      & \cellcolor{blue!15}{\textbf{1.00+-0.00}} & \cellcolor{blue!15}{\textbf{0.99+-0.01}} & \cellcolor{blue!15}{\textbf{0.99+-0.01}} & \cellcolor{blue!15}{\textbf{0.78+-0.12}} & \cellcolor{blue!15}{\textbf{0.84+-0.10}} & 0.74+-0.03 \\ 
     & & \multicolumn{1}{r|}{\cellcolor{blue!15}{\textbf{+4\%}}} & \multicolumn{1}{r|}{\cellcolor{blue!15}{\textbf{+2\%}}} & \multicolumn{1}{r|}{\cellcolor{blue!15}{\textbf{+1\%}}} &  \multicolumn{1}{r|}{\cellcolor{blue!15}{\textbf{+6\%}}} & \multicolumn{1}{r|}{\cellcolor{blue!15}{\textbf{+12\%}}} & \multicolumn{1}{r}{0\%} \\ \midrule
     
     \multirow{7}{*}{\rotatebox[origin=c]{90}{\textbf{\textit{EO}}}} 
     & (1)                & 0.57+-0.41 & 0.79+-0.30 & 0.91+-0.02 & 0.83+-0.08 & 0.89+-0.07 & 0.90+-0.02 \\ 
     & (2)                & 0.52+-0.47 & 0.73+-0.39 & 0.92+-0.03 & 0.86+-0.05 & 0.88+-0.13 & 0.90+-0.02 \\
     & (3)                & 0.92+-0.03 & 0.92+-0.03 & 0.92+-0.03 & 0.91+-0.02 & 0.91+-0.02 & 0.91+-0.02 \\
     & (4)                & 0.67+-0.46 & 0.76+-0.41 & \cellcolor{red!15}{\textbf{0.99+-0.01}} & 0.93+-0.05 & 0.91+-0.04 & 0.91+-0.02 \\  
     & (5)                & 0.86+-0.30 & 0.77+-0.41 & \cellcolor{red!15}{\textbf{0.99+-0.01}} & 0.90+-0.07 & 0.90+-0.05 & 0.91+-0.02 \\
     & (7)                & 0.62+-0.29 & 0.88+-0.12 & 0.91+-0.03 & 0.86+-0.05 & 0.90+-0.05 & 0.90+-0.02 \\
     & \textbf{(10)}      & \cellcolor{blue!15}{\textbf{0.96+-0.04}} & \cellcolor{blue!15}{\textbf{0.98+-0.01}} & 0.96+-0.01 & \cellcolor{blue!15}{\textbf{0.94+-0.04}} & \cellcolor{blue!15}{\textbf{0.99+-0.01}} & \cellcolor{blue!15}{\textbf{0.97+-0.03}} \\ 
     & & \multicolumn{1}{r|}{\cellcolor{blue!15}{\textbf{+4\%}}} & \multicolumn{1}{r|}{\cellcolor{blue!15}{\textbf{+6\%}}} & \multicolumn{1}{r|}{\cellcolor{red!15}{\textbf{-3\%}}} & \multicolumn{1}{r|}{\cellcolor{blue!15}{\textbf{+1\%}}} & \multicolumn{1}{r|}{\cellcolor{blue!15}{\textbf{+8\%}}} & \multicolumn{1}{r}{\cellcolor{blue!15}{\textbf{+6\%}}} \\ \midrule
     
     \multirow{7}{*}{\rotatebox[origin=c]{90}{\textbf{\textit{EQO}}}} 
     & (1)               & 0.50+-0.32 & 0.67+-0.29 & 0.75+-0.05 & 0.59+-0.17 & 0.62+-0.15 & 0.58+-0.02 \\ 
     & (2)               & 0.49+-0.39 & 0.62+-0.34 & 0.79+-0.08 & 0.63+-0.16 & 0.60+-0.14 & 0.58+-0.02 \\
     & (3)               & 0.78+-0.05 & 0.78+-0.05 & 0.78+-0.05 & 0.59+-0.02 & 0.59+-0.02 & 0.59+-0.02 \\
     & (4)               & 0.64+-0.44 & 0.74+-0.40 & \cellcolor{red!15}{\textbf{0.97+-0.02}} & 0.78+-0.05 & 0.80+-0.06 & 0.82+-0.03 \\  
     & (5)               & 0.83+-0.16 & 0.82+-0.19 & \cellcolor{red!15}{\textbf{0.97+-0.02}} & 0.76+-0.14 & 0.78+-0.12 & 0.82+-0.03 \\
     & (8)               & 0.64+-0.25 & 0.78+-0.12 & 0.74+-0.04 & 0.62+-0.13 & 0.73+-0.14 & 0.82+-0.03 \\
     & \textbf{(11)}     & \cellcolor{blue!15}{\textbf{0.92+-0.06}} & \cellcolor{blue!15}{\textbf{0.95+-0.04}} & 0.90+-0.04 & \cellcolor{blue!15}{\textbf{0.81+-0.06}} & \cellcolor{blue!15}{\textbf{0.89+-0.08}} & \cellcolor{blue!15}{\textbf{0.83+-0.04}} \\ 
     & & \multicolumn{1}{r|}{\cellcolor{blue!15}{\textbf{+9\%}}} & \multicolumn{1}{r|}{\cellcolor{blue!15}{\textbf{+13\%}}} & \multicolumn{1}{r|}{\cellcolor{red!15}{\textbf{-7\%}}} & \multicolumn{1}{r|}{\cellcolor{blue!15}{\textbf{+3\%}}} & \multicolumn{1}{r|}{\cellcolor{blue!15}{\textbf{+9\%}}} & \multicolumn{1}{r}{\cellcolor{blue!15}{\textbf{+1\%}}} \\ \bottomrule
\end{tabular}
\end{table}

\paragraph{Results of Last Communication Round} Tables \ref{table:res} and \ref{table:res-2} present the corresponding results of the last communication round for the \texttt{COMPAS}, the \texttt{Adult}, the \texttt{Law School} and the \texttt{Dutch} datasets \footnote{Since there are variations in the results due to different hyperparameters, we choose the outcome that achieves a good balance between accuracy and fairness, but prioritizes fairness by allowing a small sacrifice in accuracy if necessary.}. Regarding the \texttt{COMPAS} dataset, FAIR-FATE was able to surpass the baselines 78\% of the times. In particular, for $\sigma=0.5$, FAIR-FATE presented substantial improvements: 7\% for $SP$, 11\% for $EO$, and 16\% for $EQO$. Concerning the \texttt{Adult} dataset, FAIR-FATE was always able to surpass the baselines with an average improvement of fairness of 10\%. In particular, for $\sigma=0.5$, FAIR-FATE presented substantial improvements: 8\% for $SP$, 7\% for $EO$, and 19\% for $EQO$. Regarding the \texttt{Law School} and the \texttt{Dutch} datasets, FAIR-FATE surpassed the baselines 78\% and 88\% of the times, respectively. Overall, the baselines only surpassed FAIR-FATE for random train-test splits (RND), which do not represent typical Federated Learning settings. 

\begin{figure}[h!]
    \centering
    \includegraphics[width=\textwidth]{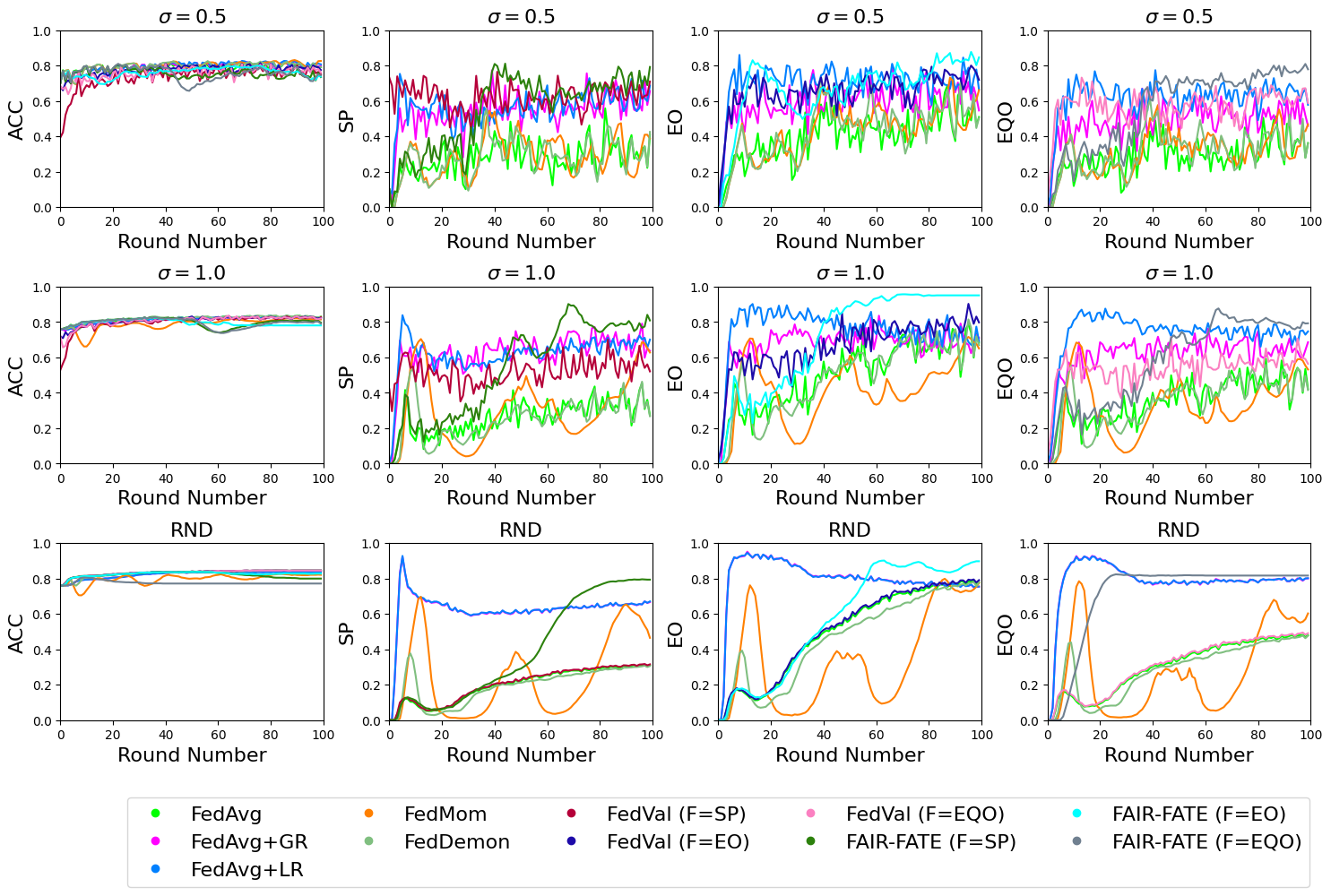}
    \caption{Performance over communication rounds on the \texttt{Adult} dataset for the different algorithms under different $\sigma$ heterogeneity levels and random (RND) splits.}
    \label{fig:rounds-1}%
\end{figure}

\begin{figure}[h!]
    \centering
    \includegraphics[width=6.5cm]{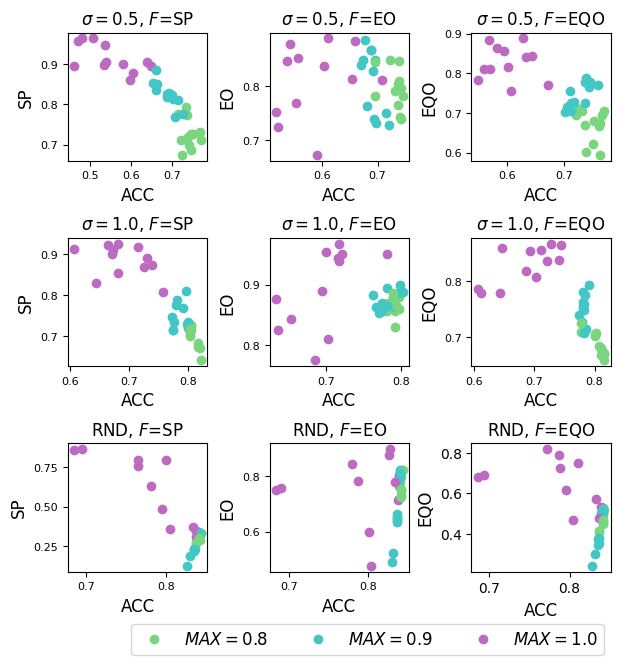}
    \caption{FAIR-FATE's fairness-accuracy trade-off on the \texttt{Adult} dataset considering $MAX$ under different $\sigma$ heterogeneity levels and random (RND) splits.}
    \label{fig:pareto-adult-max}%
\end{figure}

\paragraph{Stability and Convergence} Figure \ref{fig:rounds-1} presents the results over communication rounds under different sensitive attribute distributions on the \texttt{Adult} dataset. It can be observed that FAIR-FATE is able to surpass the baselines in terms of fairness, especially in heterogeneous configurations ($\sigma=0.5$ and $\sigma=1.0$), without neglecting the classification performance. Moreover, the stability of the algorithms decreases in the presence of heterogeneous data. However, FAIR-FATE is able to overcome the oscillations of non-fair gradients using a fairness-aware Momentum term and, consequently, the fairness performance does not drop after some communication rounds.

\paragraph{Fairness vs Model Performance Trade-off using $MAX$} We further investigate the fairness and model performance trade-off of FAIR-FATE, controlled by the $MAX$ hyperparameter. Figure \ref{fig:pareto-adult-max} present FAIR-FATE's fairness-accuracy trade-offs considering the $MAX$ hyperparameter on the \texttt{Adult} dataset. It can be observed that generally $MAX$ can be used to select different trade-offs, with higher values of $MAX$ resulting in higher values of fairness and lower values of accuracy, and lower values of $MAX$ resulting in lower values of fairness and higher values of accuracy. On a broad level, it can be observed that the results produce a shape similar to a pareto front, especially for $F=SP$. However, for $F=EO$ and $F=EQO$ this phenomenon is not as noticeable. This might be justified by the fact that these metrics consider the actual predictions ($\hat{Y}$), and, consequently, are more aligned with the performance of the algorithm. 

\paragraph{Effect of $\rho$, $\lambda_0$ and $\beta_0$} Experimental results suggest that the values of $\beta_0$, $\lambda_0$, and $\rho$ should be chosen based on experimentation. Lower values of $\rho$ and $\lambda_0$ could be preferable in cases where there is significant heterogeneity of the data with respect to the sensitive attributes. These lower values allow the global model to adjust to the diversity in the data more slowly and capture the complexity of the data. Regarding the $\beta_0$ hyperparameter, it is more difficult to analyse its impact. As a result, we recommend tuning these hyperparameters to achieve the best results.

\section{Conclusion and Future Work}

We proposed FAIR-FATE: a fairness-aware Federated Learning algorithm that uses a novel aggregation technique that estimates a fair Momentum update based on the fairness of each client participating in the federation. In addition, FAIR-FATE does not need clients' restricted information, respecting the data privacy constraints associated with Federated Learning. Furthermore, FAIR-FATE is a flexible algorithm that does not rely on any local fairness strategy at each client. 

To the best of our knowledge, this is the first approach in machine learning that aims to achieve fairness using a fair Momentum estimate. Experimental results on real-world datasets demonstrate that FAIR-FATE can effectively increase the fairness results under different data distributions compared to state-of-the-art fair Federated Learning approaches.

Interesting directions for future work are: testing FAIR-FATE using multiple fairness metrics and sensitive attributes; extending the work to various application scenarios such as clustering; focusing on individual fairness notions.

\section*{Acknowledgments}

This work is funded by the FCT - Foundation for Science and Technology, I.P./MCTES through national funds (PIDDAC), within the scope of CISUC R\&D Unit - UIDB/00326/2020 or project code UIDP/00326/2020. This work was supported in part by the Portuguese Foundation for Science and Technology (FCT) Research Grants 2021.05763.BD.

\bibliographystyle{splncs04}
\bibliography{bib.bib}

\begin{thebibliography}{10}
\providecommand{\url}[1]{\texttt{#1}}
\providecommand{\urlprefix}{URL }
\providecommand{\doi}[1]{https://doi.org/#1}

\bibitem{mitigating-bias-fl}
Abay, A., Zhou, Y., Baracaldo, N., Rajamoni, S., Chuba, E., Ludwig, H.:
  Mitigating bias in federated learning. arXiv preprint arXiv:2012.02447
  (2020)

\bibitem{compas}
Angwin, J., Larson, J., Mattu, S., Kirchner, L.: {Machine Bias}  (2016),
  \url{https://www.propublica.org/article/machine-bias-risk-assessments-in-criminal-sentencing}

\bibitem{adult}
Asuncion, A., Newman., D.: Adult data set. In: UCI Machine Learning Repository
  (2017), \url{https://archive.ics.uci.edu/ml/datasets/adult}

\bibitem{validation-set}
Bhagoji, A.N., Chakraborty, S., Mittal, P., Calo, S.: Analyzing federated
  learning through an adversarial lens. In: Chaudhuri, K., Salakhutdinov, R.
  (eds.) Proceedings of the 36th International Conference on Machine Learning.
  Proceedings of Machine Learning Research, vol.~97, pp. 634--643. PMLR (09--15
  Jun 2019)

\bibitem{survey_fairness_ml}
Caton, S., Haas, C.: Fairness in machine learning: A survey. arXiv preprint
  arXiv:2010.04053  (2020)

\bibitem{chen2022demon}
Chen, J., Wolfe, C., Li, Z., Kyrillidis, A.: Demon: improved neural network
  training with momentum decay. In: ICASSP 2022-2022 IEEE International
  Conference on Acoustics, Speech and Signal Processing (ICASSP). pp.
  3958--3962. IEEE (2022)

\bibitem{parecido-group-fairness-fl}
Ezzeldin, Y.H., Yan, S., He, C., Ferrara, E., Avestimehr, S.: Fairfed: Enabling
  group fairness in federated learning. arXiv preprint arXiv:2110.00857  (2021)

\bibitem{best-fl}
Fernandes, M., Silva, C., Arrais, J.P., Cardoso, A., Ribeiro, B.: Decay
  momentum for improving federated learning. ESANN 2021 pp. 17--22 (2021).
  \doi{10.14428/esann/2021.ES2021-106}

\bibitem{equality-of-opportunity}
Hardt, M., Price, E., Srebro, N.: Equality of opportunity in supervised
  learning. Advances in neural information processing systems  \textbf{29}
  (2016)

\bibitem{heo2020adamp}
Heo, B., Chun, S., Oh, S.J., Han, D., Yun, S., Kim, G., Uh, Y., Ha, J.W.:
  Adamp: Slowing down the slowdown for momentum optimizers on scale-invariant
  weights. arXiv preprint arXiv:2006.08217  (2020)

\bibitem{distributions-nonidd}
Hsu, T.M.H., Qi, H., Brown, M.: Measuring the effects of non-identical data
  distribution for federated visual classification. arXiv preprint
  arXiv:1909.06335  (2019)

\bibitem{parecido-fairl-fl-het-face}
Kanaparthy, S., Padala, M., Damle, S., Gujar, S.: Fair federated learning for
  heterogeneous face data. arXiv preprint arXiv:2109.02351  (2021)

\bibitem{dutch-census}
Van~der Laan, P.: The 2001 census in the netherlands. In: Conference The Census
  of Population (2000)

\bibitem{accelerating_fl_momentum}
Liu, W., Chen, L., Chen, Y., Zhang, W.: Accelerating federated learning via
  momentum gradient descent. IEEE Transactions on Parallel and Distributed
  Systems  \textbf{PP}, ~1--1 (02 2020). \doi{10.1109/TPDS.2020.2975189}

\bibitem{Federated_Learning_Avg}
McMahan, H., Moore, E., Ramage, D., Hampson, S., y~Arcass, B.A.:
  Communication-efficient learning of deep networks from decentralized data.
  In: In AAAI Fall Symposium. Google, Inc., 651 N 34th St., Seattle, WA 98103
  USA (2017)

\bibitem{parecido-mehrabi}
Mehrabi, N., de~Lichy, C., McKay, J., He, C., Campbell, W.: Towards
  multi-objective statistically fair federated learning. arXiv preprint
  arXiv:2201.09917  (2022)

\bibitem{momentumgd}
Qian, N.: On the momentum term in gradient descent learning algorithms. Neural
  Networks  \textbf{12}(1),  145 -- 151 (1999)

\bibitem{best-fairness}
Salazar, T., Santos, M.S., Araujo, H., Abreu, P.H.: Fawos: Fairness-aware
  oversampling algorithm based on distributions of sensitive attributes. IEEE
  Access  \textbf{9},  81370--81379 (2021). \doi{10.1109/ACCESS.2021.3084121}

\bibitem{law-school}
Wightman, L.: Lsac national longitudinal bar passage study. In: lsac research
  report series (1998)

\bibitem{zafar-2017}
Zafar, M.B., Valera, I., Rogriguez, M.G., Gummadi, K.P.: Fairness constraints:
  Mechanisms for fair classification. In: Artificial Intelligence and
  Statistics. pp. 962--970. PMLR (2017)

\end{thebibliography}

\end{document}